\documentclass{article}

\usepackage{PRIMEarxiv}

\usepackage[utf8]{inputenc} % allow utf-8 input
\usepackage[T1]{fontenc}    % use 8-bit T1 fonts
\usepackage{hyperref}       % hyperlinks
\usepackage{url}            % simple URL typesetting
\usepackage{booktabs}       % professional-quality tables
\usepackage{amsfonts}       % blackboard math symbols
\usepackage{nicefrac}       % compact symbols for 1/2, etc.
\usepackage{microtype}      % microtypography
\usepackage{lipsum}
\usepackage{fancyhdr}       % header
\usepackage{graphicx}       % graphics
\graphicspath{{media/}}     % organize your images and other figures under media/ folder
\usepackage{graphicx}
\usepackage{subcaption}
\usepackage{xcolor}
\usepackage{url}
\newcommand{\our}{RIECNN}
\newcommand{\ourb}{\textbf{RIECNN}}

\title{RIECNN: Real-Time Image Enhanced CNN for Traffic Sign Recognition
\thanks{\textit{\underline{Citation}}:
\textbf{Abdel-Salam, R., Mostafa, R. \& Abdel-Gawad, A.H.}
RIECNN: real-time image enhanced CNN for traffic sign recognition.
Neural Computing \& Applications, 34, 6085--6096 (2022).
DOI: \href{https://doi.org/10.1007/s00521-021-06762-5}{10.1007/s00521-021-06762-5}.}
}

\author{
  Reem Abdel-Salam \\
  Computer Engineering Department \\
  Cairo University \\
  Giza, Egypt\\
  \texttt{reem.abdelsalam13@gmail.com} \\
  \And
  Rana Mostafa \\
  Computer Engineering Department \\
  Cairo University \\
  Giza, Egypt\\
  \texttt{ranamostafamohsen96@yahoo.com} \\
  \And
  Ahmed H. Abdel-Gawad \\
  Computer Engineering Department \\
  Cairo University \\
  Giza, Egypt\\
  \texttt{ahamdy@eng.cu.edu.eg} \\
}

\begin{document}
\maketitle

\begin{abstract}
Traffic sign recognition plays a crucial role in the development of autonomous cars to reduce the accident rate and promote road safety. It has been a necessity to address traffic signs that are affected significantly by the environment as well as poor real-time performance for deep-learning state-of-the-art algorithms. In this paper, we introduce Real-Time Image Enhanced CNN ({\our}) for Traffic Sign Recognition. {\our} is a real-time, novel approach that tackles multiple, diverse traffic sign datasets, and out-performs the state-of-the-art architectures in terms of recognition rate and execution time. Experiments are conducted using the German Traffic Sign Benchmark (GTSRB), the Belgium Traffic Sign Classification (BTSC), and the Croatian Traffic Sign (rMASTIF) benchmark. Experimental results show that our approach has achieved the highest recognition rate for all Benchmarks, achieving a recognition accuracy of 99.75\% for GTSRB, 99.25\% for BTSC and 99.55\% for rMASTIF. In terms of latency and meeting the real-time constraint, the pre-processing time and inference time together do not exceed 1.3 ms per image. Not only have our proposed approach achieved remarkably high accuracy with real-time performance, but it also demonstrated robustness against traffic sign recognition challenges such as brightness and contrast variations in the environment.\end{abstract}

% keywords can be removed

\section{Introduction}
	\label{intro}
	Traffic sign recognition is crucial in the self-driving field for the automotive industry. Self-driving technology can assist, or even independently complete the driving operation, which plays a remarkable role to facilitate driving and considerably reduces the risk of accidents. With the development of the technology, it became a necessity to promote smart cars that can recognize traffic signs in an actual environment in real-time.
	
	Smart cars are installed with cameras that capture road traffic images and analyze in real-time. The target of traffic sign recognition is to highlight interesting traffic sign regions and classify the type of traffic sign effectively. The challenge is the large variance in the captured images quality in accordance with the various environmental conditions. The quality of the image can be affected by several reasons: a) the camera quality in terms of image resolution, b) the brightness of the image captured which can be too bright, too dim, or contains spotlights, c) the weather condition at the time of image capture which can be snowy, rainy, or foggy, d) the motion blurriness due to the photo being taken at high speed in a way that the traffic sign is barely recognized, e) the road situations that can obscure parts of the traffic sign during the image capture. The uncontrolled environment aspect of the traffic sign recognition problem is the main source of the above challenges \cite{SegmentedBased2018}.
	
	Although the deep learning-based traffic sign recognition algorithms achieve high recognition accuracy, these algorithms are highly complex and suffer from long processing time. Moreover, limitations arise due to the high system requirements and the complex structures of training models \cite{Jose2018}. Therefore, further improvements of the traffic sign recognition algorithm are needed. The complexity of the deep neural network architecture is mainly to handle the image problems (the uncontrolled aspect). However, alleviating the image problems can be better tackled using image enhancement techniques; thus,  by it achieves better accuracy while simplifying the network complexity.
	
	In this work, we propose an improved traffic sign recognition algorithm for intelligent vehicles. It utilizes image enhancements to correct image problems since it is more efficient in both factors: accuracy and speed. It achieves higher accuracy while reducing the model size; thus achieving faster inference. The time overhead of the image enhancement is very low relative to the inference time of the deep neural network as it is performed in very low resolution (i.e., $60\times60$). First, captured images are pre-processed and then followed by a deep learning model. Experiments are conducted using multiple traffic sign benchmarks: GTSRB, BTSC, and rMASTIF to prove the generalization of our approach.
	
	The main contributions of this paper are:
	\begin{enumerate}
		\item {\our}, a compact image enhanced CNN model for traffic sign recognition, is proposed that targets higher accuracy, yet faster inference than previous techniques.
		\item The four different image enhancement stages were implemented and assessed for their impact in enhancing the accuracy of the CNN model whether separately or in combination together. This can give general insights into which image enhancement techniques are effective in boosting the performance of CNN models for other domains with uncontrolled environments.
		\item The accuracy of {\our} was assessed by applying it on GTSRB, BTSC, and rMASTIF achieving the highest recognition accuracy in all three datasets compared to all published work.
	\end{enumerate}
	
	The remainder of the paper is organized as follows: In Section 2, an overview of the related work is presented. In Section 3, our proposed approach is described in detail. In Section 4, the experiments on traffic sign recognition datasets are analyzed, and the performance of state-of-the-art architectures against {\our} are compared  \footnote{The source code for the developed models and experiments can be
		found through: \url{https://github.com/RanaMostafaAbdElMohsen/Traffic_Sign_Recognition}}. In Section 5, the conclusion and recommendations for future work are outlined.
	
	\section{Related Work}
	Ciresan et al., \textit{Multi-column deep neural network for traffic sign classification} \cite{MCDNN}, introduced a Multi-Column DNN that utilizes a committee of CNNs. The authors used 25 different CNN. Each CNN is trained on different pre-processed data, achieving an accuracy of 99.46\%. Since the GTSRB dataset suffers from high contrast variation, the authors used several pre-processing steps to overcome this such as: image adjustment, histogram equalization, adaptive histogram equalization, and contrast normalization. The real-time performance of this approach is poor due to its large number of parameters ($\sim$ 90 M). García et al., \textit{Deep neural network for traffic sign recognition systems: An analysis of spatial transformers and stochastic optimisation methods} \cite{STDNN}, presented a deep learning architecture with the use of the Spatial Transformers network. Spatial transformer layers, \textit{CNN with 3 Spatial Transformers} \cite{SpatialTransfromener}, are used on feature layer maps to perform explicit geometric transformations to concentrate on the object to be learned, gradually eliminating background and geometrical noise. The network included three spatial transformer layers with 14 Million parameters. For processing the dataset, global normalization and local contrast normalization with Gaussian kernels were computed in order to enhance the edges in the images. Experiments on this architecture were conducted using GTSRB and BTSC with an achieved recognition rate of 99.71\% and 98.86\% respectively.
	
	Sermanet et al., \textit{Traffic sign recognition with multi-scale Convolutional Networks} \cite{MSCNN}, introduced a multi-scale CNN where the outputs of all stages contribute to the classification by feeding them to the classifier. This approach achieved 98.31\% concerning the IJCNN 2011 session competition using GTSRB Benchmark.  Then, the authors increased the model's depth in terms of layers and ignored color information, which achieved a 99.17\% recognition accuracy. Saha et al., \textit{Total Recall: Understanding Traffic Signs using Deep Hierarchical Convolutional Neural Networks} \cite{DHCNN}, presented a deep hierarchical residual CNN model with a dilated skip connection. The number of the model parameters reported was 6.256 M. This technique achieved 99.33\% and 99.17\% recognition accuracy in GTSRB and BTSC datasets respectively.  
	\begin{figure}[ht]
		\centering
		\includegraphics[width=\linewidth]{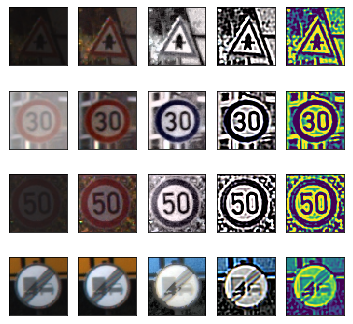}
		\caption{Sample of GTSRB traffic signs after pre-processing, left images are original unprocessed ones and the next images to the right display the result after each stage of pre-processing in order: 1) Image contrast enhancement, 2) Retinex algorithm, 3) Histogram equalization, and 4) Edge enhancement with gray-scale conversion}
		\label{fig:SampleGTSRBPre-Processed}
	\end{figure} 
	
	\begin{figure*}[ht]
		\centering
		\includegraphics[width=0.7\linewidth]{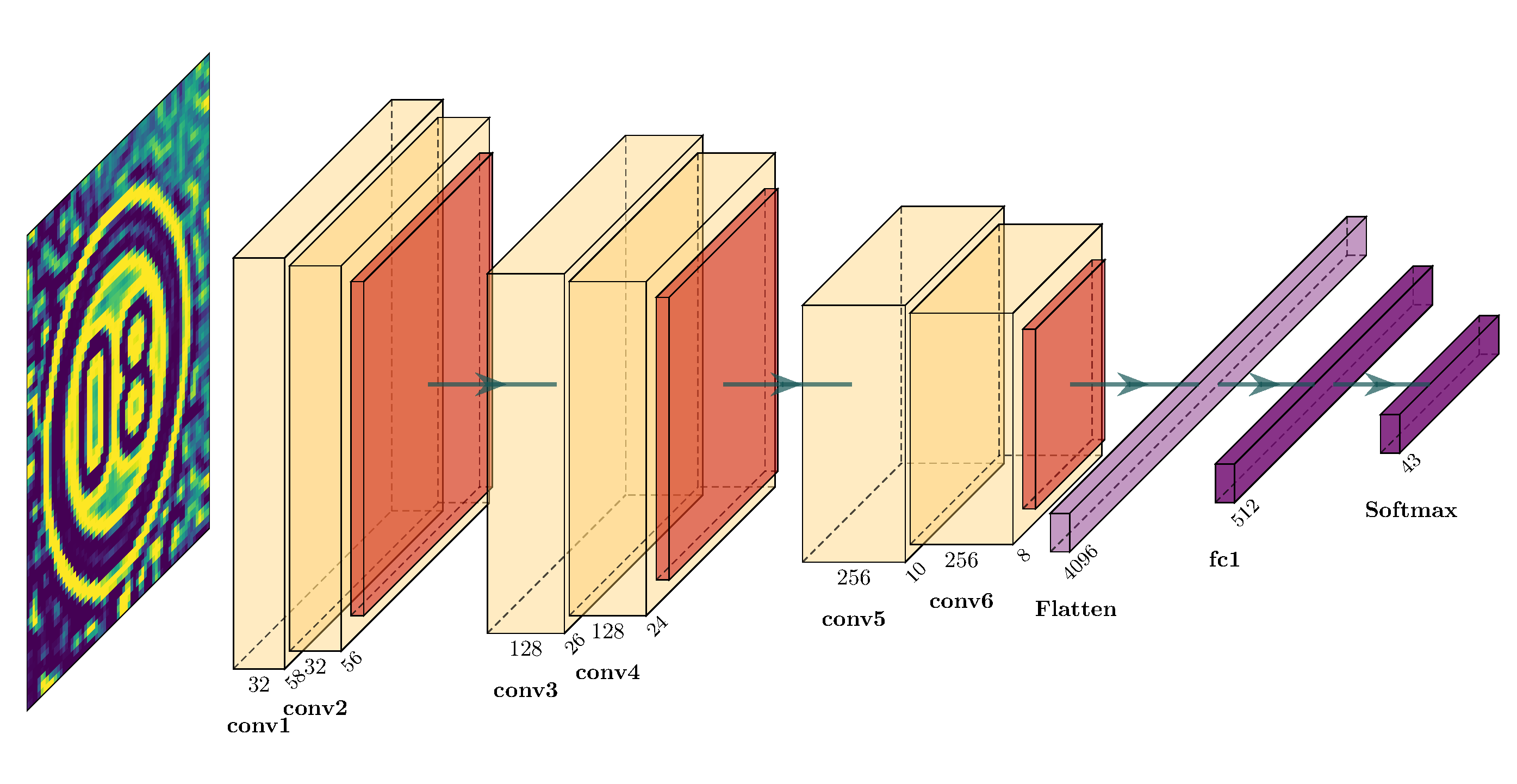}
		\caption{Our Proposed CNN Architecture}
		\label{fig:RIECNNVisualization}
	\end{figure*} 
	
	\begin{figure}[ht]
		\centering
		\includegraphics[width=0.3\linewidth]{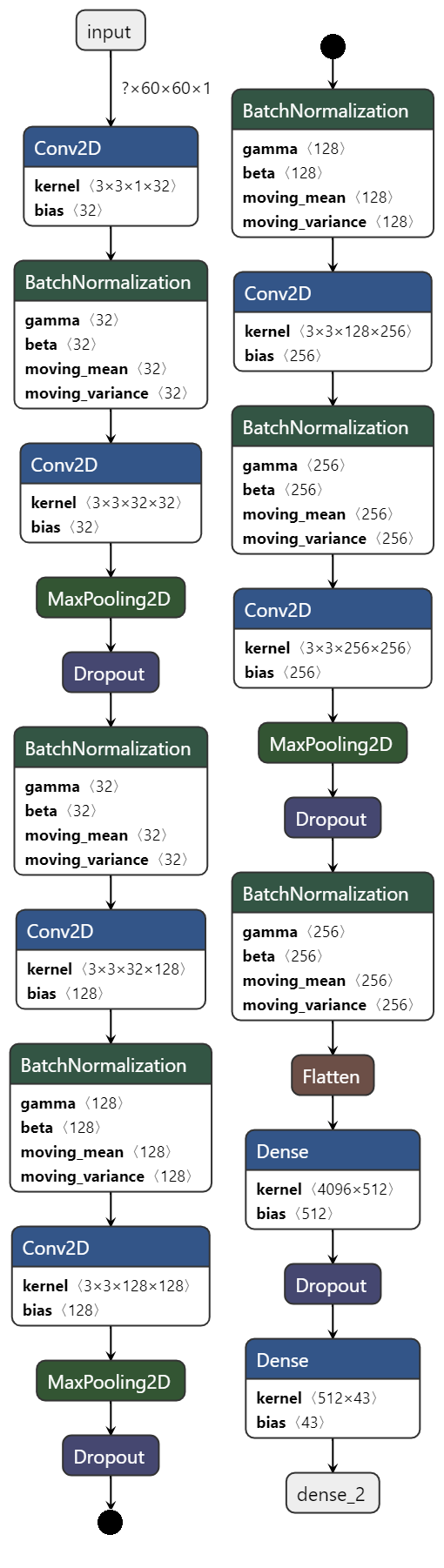}
		\caption{Full model details}
		\label{fig:ModelParams}
	\end{figure}
	Zhang et al., \textit{Lightweight deep network for traffic sign classification} \cite{Zhang2019LightweightDN}, proposed a lightweight DNN model using the knowledge distillation technique. A teacher network is first trained as usual on the given dataset. The authors improved the teacher model by adding a module that combines the stream of the feature with a dense layer. Then, a shallow network "student network" is trained through the softened output of the teacher network on the target datasets. The student network with 0.8 million parameters {whilst the teacher network with 7.4 million parameters} achieved optimal 99.61\% and 99.13\% recognition accuracy on the GTSRB and BTSC respectively. {This approach is trained for 300 epochs}. {Although it achieves a competitive accuracy with a low number of parameters, it suffers from convergence issues. The convergence issues arise since the student model is trained by minimizing loss function between it and the teacher model.} {It depends on how well the teacher is trained in order for the student to train well}. Therefore, one needs to make sure that the student models are learning representative features. Also, for each dataset, a teacher model must be trained.Thus, it may need many experiments with training the student model, due to instability issues, to be able to achieve high accuracy. 
	
	Mao et al., \textit{Hierarchical CNN for traffic sign recognition} \cite{Mao2016HierarchicalCF}, proposed a Hierarchical CNN (HCNN) model, which is inspired by coarse-to-fine human learning. To begin with, the dataset is divided into K-subsets. HCNN algorithm trains each subset by a single CNN. The HCNN approach achieves 99.67\% accuracy on the GTSRB dataset. This approach may be rather time-consuming when applying this approach to other datasets to determine the optimal number of K-subsets.
	
	Jurišić et al., \textit{Multiple-dataset Traffic Sign Classification with OneCNN} \cite{OneCNN}, proposed a deep CNN model with a drop-out layer, and experiments were conducted on multiple datasets: GTSRB, BTSC, and rMASTIF. To increase the size of the datasets while training, the authors augmented the datasets by duplicating the images and altering these duplicates. For duplicate images, histogram equalization and 10\% padding were applied. The proposed model was trained using 25K epochs. Experiments have shown that it achieved a competitive accuracy, out-performing human accuracy, and other architectures. Zeng et al., \textit{Traffic Sign Recognition Using Deep Convolutional Networks and Extreme Learning Machine}  \cite{Zeng2015TrafficSR}, proposed a CNN-ELM (extreme learning machine) model, which combines CNN's ability to extract features and ELM ability of high generalization as a classifier. The CNN-ELM network achieves a 99.40 \% accuracy on the GTSRB dataset. Before feeding an image to the model, its average is subtracted to ensure illumination invariance.
	This accuracy is achieved without any data augmentation.
	
	The challenge still exists to introduce a light-weight, stable approach that can be deployed in automotive systems, and at the same time,  yields high competitive accuracy on multiple, diverse datasets.
	
	\section{Methodology}
	{\our} relies on two stages to classify traffic signs universally for any dataset: the pre-processing stage and the deep learning architecture stage. 
	\subsection{Pre-processing}
	\label{Pre-Processing}
	Not only are there environmental challenges as mentioned above, but there are challenges related to producing a model with a low number of parameters that can be deployed on autonomous driving systems. To overcome these challenges,  pre-processing and image enhancement techniques are the best choice to achieve the best reduction of image problems. The other approaches that rely on increasing the depth of neural network models are not as efficient in both aspects; recognition accuracy and speed. During the pre-processing stage, all images (RGB) are resized to $60\times60$. To begin with, image contrast enhancement \cite{Ying2017ANI} is applied to improve poor quality images. Then, it is followed by the Multi-scale Retinex algorithm \cite{Jobson1997AMR} that is used to improve color consistency. Histogram equalization and edge enhancement are then applied to further enhance the contrast and sharpen the edges. Finally, all images are converted to gray-scale.
	
	A considerable number of traffic sign datasets suffer from low-resolution issues such as blurriness and brightness. Therefore, for better recognition, the image color quality is in need to be enhanced in order to be more color consistent and restore lost information that is due to brightness issues. We applied a Multi-scale Retinex algorithm that achieves simultaneous color consistency rendition. To begin with, Multi-scale Retinex algorithm \cite{Jobson1997AMR} transforms the image into another space \textit{aperture mode}. The aperture mode makes image pixels less dependent on illumination distribution.
	The following equation outlines how to generate aperture mode image,
	
	$$ I_i^{'}(x,y)= I_i(x,y)/ \sum_{i=1}^{S} I_i(x,y)$$
	
	where the $I_i{'}$ is aperture mode image and $I_i$ is the original image.
	
	In order to enhance the image quality, after the space transformation, non-linear transformation is further applied to each pixel,
	
	$$ C_i(x,y)=\beta \log [\alpha  I_i^{'}(x,y)]$$ 
	
	where $C_i$ is the final output image, $\beta$ is the gain constant, and $\alpha$ is the strength of the non-linearity. Unfortunately, the Multi-scale Retinex algorithm suffers from a halo-like artifact in the high contrast region images. It does not work well with images that suffer from high brightness. 
	
	In order for the Retinex algorithm to work correctly, we need to deal with traffic sign images that suffer from reduced-brightness or over-brightness issues. Image Contrast Enhancement Algorithm is first applied before the Retinex Algorithm to fix low-contrast and high-contrast images \cite{Ying2017ANI}. This technique overcomes the Multi-scale Retinex algorithm weakness. The algorithm attempts to find the optimal exposure ratio for an image. It first calculates the weight matrix for each pixel in which small values indicate under-exposed pixels; meanwhile, large values indicate well-exposed pixels. The synthetic image is a multiplication of the beta-gamma corrected original image with the inverse of the weight matrix. The resultant image is a combination of both the weight matrix multiplied with the original image and the synthetic image. This methodology is achieved by the following steps. It first applies brightening over the whole original image. Then, it adds another version of the image where the low contrast regions are high and high contrast regions are low. This acts as weight blending for each pixel; thus, this contributes to maintain a reasonable lightening effect throughout the whole image. Shortly, the weight matrix is calculated. Then, the synthetic multi-exposure images are introduced by applying beta-gamma correction where the synthetic image is exposed better in the regions where the original image under-exposed. Finally, the resultant image is obtained by fusing the input and the synthetic image using the weight matrix as the equation below,
	$$ R = W * P + (1-W) * P^{*} $$ 
	where $R$ is the resultant image, $W$ is the weight matrix, $P$ is the input image, and $P^{*}$ is the synthetic image.

	\begin{table}[h]
		\caption{{Different pre-processing techniques used to overcome GTSRB, BTSC, and rMASTIF challenges}}
		\centering
		\begin{tabular}{|p{2.5cm}|p{4.5cm}|} 
			\hline
				{Model Name} &	{Pre-processing technique}\\ 
			\hline
			 	{\our}& \begin{itemize}
                      \item 	{Retinex algorithm}
                      \item 	{Contrast Enhancement}
                      \item 	{Edge enhancement}
                      \item 	{Histogram equalization}
                      \end{itemize} \\ \hline
          	{MCDNN} & 
                      \begin{itemize}
                      \item {Image Adjustment}
                      \item {Histogram Equalization}
                      \item {Adaptive Histogram Equalization}
                      \item {Contrast Normalization}
                      \end{itemize} \\ \hline
                      
          	{STDNN} &     \begin{itemize}
                      \item {Global normalisation} 
                      \item {Local contrast normalisation}
                      \end{itemize} \\ \hline

          	{MSCNN} &     \begin{itemize}
                      \item {Global normalisation} 
                      \item {Local normalization}
                      \end{itemize} \\ \hline
                      
          	{LWDNN} &     	{-} \\ \hline
           	{OneCNN} &     {-} \\ \hline
           	{DHCNN} &     {-} \\ \hline
                      
			\hline
		\end{tabular}
		\label{Differentpreprocessingtechniques}
	\end{table}

	{Majority of the traffic sign datasets suffer mainly from blurriness, aging effects, and lightning effects. Prior work used pre-processing techniques to overcome theses challenges as shown in table \ref{Differentpreprocessingtechniques}.
	To overcome these challenges, we used mainly Retinex algorithm and Image  Contrast  Enhancement. Image contrast enhancement is mainly used to tackle under and over-enhancement problems in images (lightning effects), while Retinex algorithm is used for the aging, lightning effects. Unfortunately, the Retinex algorithm suffers from halo-like artifacts in the high contrast region images, which could worsen image quality. Therefore, we applied image contrast enhancement first. By applying these two algorithms, we get less contrast and lightness distorted images. \\
	STDNN, and MSCNN uses global contrast normalization  and local contrast normalization. Global contrast normalization  subtracts each pixel value of image by the mean and divides it by the standard deviation. It aims to prevent images from having varying amounts of contrast. However, images with very low, but non-zero contrast, often contain less information. Local contrast normalization performs local subtraction and division normalization, enforcing a sort of local competition between adjacent features in a feature map, and between features at the same spatial location in different feature maps, which may lead to a whitening effect in images. Therefore, the main disadvantages of global and local contrast normalization are having less features in images and whitening effects (distorted image).\\}
	
	Our approach has effectively enhanced the quality of captured images that suffer from poor contrast, reduced brightness, and low resolution. Figure \ref{fig:SampleGTSRBPre-Processed} shows a sample of GTSRB traffic signs in different environmental conditions and their output after each stage of pre-processing in order. It can be inferred that our pre-processing approach has proven to enhance a set of diverse, low-quality image conditions: extreme darkness, low resolution, and blurry frames. It was also noticed that our approach even improves good resolution images to be sharper and visually enhanced.
	We effectively treated other problems such as blurriness, perspective, and occlusion by utilizing data augmentation.
	
	\subsection{Deep Learning Architecture}
	\label{RIECNNArch}
	Inspired by the VGG16 network model \cite{VGG16Paper} that was proposed in 2014, we designed a similar architecture yet shallower in depth. The VGG16 model consists of five convolutional blocks. Each of the first two blocks consists of two convolutional layers with the same parameters followed by a max pool. Each of the third, fourth, and fifth blocks consists of three convolutional layers followed by a max pool. Finally, the convolutional blocks are followed by three fully connected layers. The VGG16 model suffers from training convergence issues, and the efficiency of recognition of the network decreases drastically.
	Our proposed CNN architecture poses a similar structure to VGG16. It uses blocks where each block consists of two convolutional layers with exact parameters followed by the max pool, etc. On the other hand, our model is 8 layers in depth with fewer number of parameters and converges better during training.
	
	Our proposed architecture, as shown in figure \ref{fig:RIECNNVisualization}, consists of successions of convolution layers, max-pooling, and batch normalization. Our architecture consists mainly of three convolutional blocks. The first block consists of 2 convolutional layers with 32 filters, each of kernel size (3x3) with a kernel regularizer followed by a batch normalization, and a dropout of 0.2. Then, the second block consists of 2 convolutional layers with 128 filters, each of kernel size (3x3) with a kernel regularizer followed by a batch normalization, and a dropout of 0.2. Finally, it is followed by two convolutional layers with 256 filters, each of kernel size (3x3) followed by a batch normalization, and a dropout of 0.2. Finally, there are two fully connected layers: the first 512-fully connected layer, followed by a dropout of 0.4, and the final output layer corresponds to the number of classes in the dataset. Full model details are shown in figure \ref{fig:ModelParams}.
	
	\begin{table}[h]
		\caption{Traffic sign datasets used for evaluation}
		\centering
		\begin{tabular}{|c|p{1.0cm}|p{1.1cm}|p{1.1cm}|p{1.0cm}|} 
			\hline
			Dataset & Total Images & Training Images & Testing Images & Number of Classes\\ 
			\hline
			GTSRB & \multicolumn{1}{|r|}{51,839} & \multicolumn{1}{|r|}{39,209} & \multicolumn{1}{|r|}{12,630} & \multicolumn{1}{|r|}{43} \\
			\hline
			BTSC & \multicolumn{1}{|r|}{7,095} & \multicolumn{1}{|r|}{4,575} & \multicolumn{1}{|r|}{2,520} & \multicolumn{1}{|r|}{62} \\
			\hline
			rMASTIF & \multicolumn{1}{|r|}{5,828} & \multicolumn{1}{|r|}{4,044} & \multicolumn{1}{|r|}{1,784} & \multicolumn{1}{|r|}{31} \\
			\hline
		\end{tabular}
		\label{DifferentDataSets}
	\end{table}
	
	\section{Experimental Results}
	\subsection{Dataset Preparation}
	
	\begin{figure}[h]
		\centering
		\includegraphics[width=\linewidth]{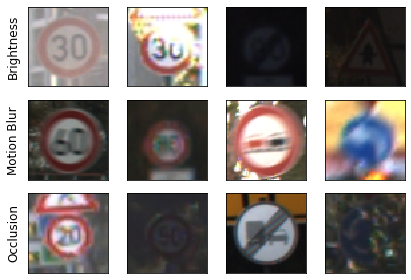}
		\caption{GTSRB images variations}
		\label{fig:GTSRB}
	\end{figure} 
	
	\begin{figure}[h]
		\centering
		\includegraphics[width=1.0\linewidth]{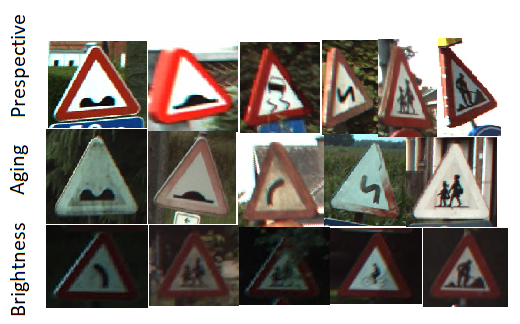}
		\caption{BTSC images variations}
		\label{fig:belgiumIssues}
	\end{figure} 
	
	\begin{figure}[h]
		\centering
		\includegraphics[width=1.0\linewidth]{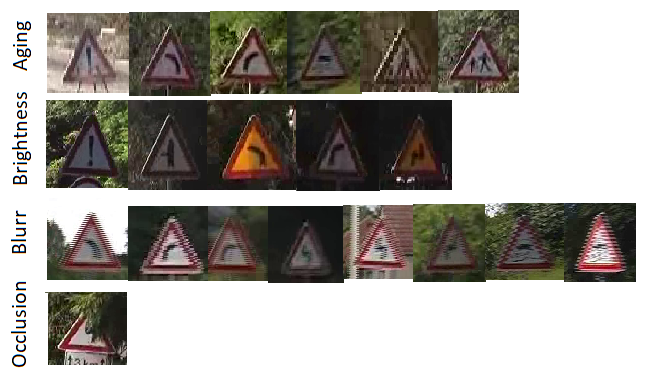}
		\caption{rMASTIF images variations}
		\label{fig:rMIssues}
	\end{figure}

	\begin{figure}[h]
		\centering
		\includegraphics[width=\linewidth]{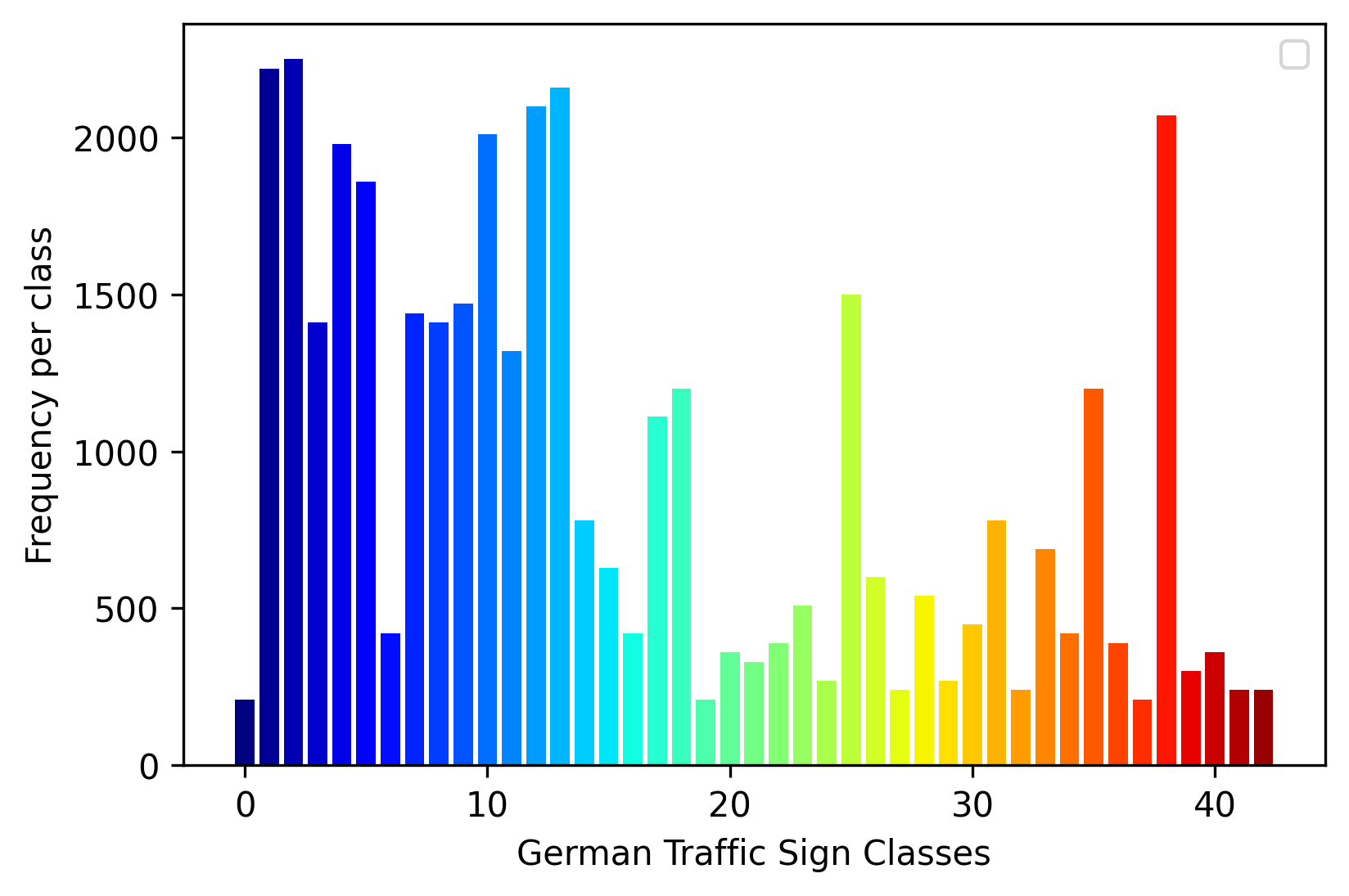}
		\caption{GTSRB dataset classes distribution}
		\label{fig:GTSRBdistribution}
	\end{figure}

	\begin{figure}[h]
		\centering
		\includegraphics[width=\linewidth]{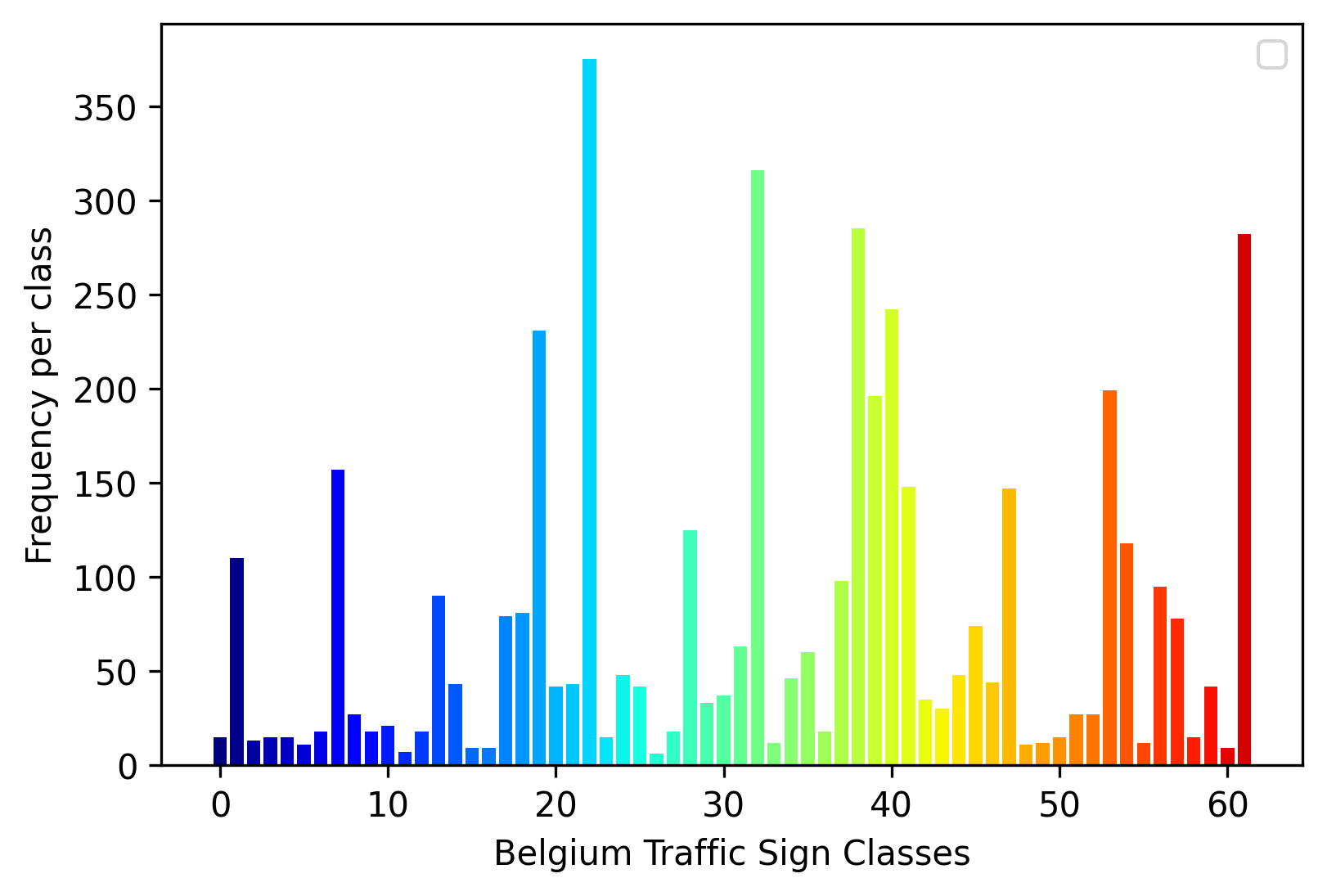}
		\caption{BTSC dataset classes distribution}
		\label{fig:BTSCdirstibution}
	\end{figure}
	
	{We conducted our experiments and compared our results with the state of the art papers using multiple benchmarks: the German Traffic Sign Recognition Benchmark (GTSRB), the Belgium Traffic Sign Classification (BTSC), and the Croatian traffic sign (rMASTIF).}
	
	Table \ref{DifferentDataSets} shows the details of the three traffic sign benchmarks used for experimental evaluation in terms of the number of total images, the number of training images, the number of test images, and the number of traffic sign classes in each dataset.
	
	The German traffic sign recognition dataset (GTSRB) consists of 39,209 training colored images, and 12,630 colored images in the test set. The GTSRB is divided into 43 traffic sign classes. The images vary in pixel size, ranging from 15x15 to 250x250 pixels. The GTSRB dataset poses a lot of challenges due to the image conditions. The images in the dataset deal with occlusions, different lighting conditions, motion blur, and perspectives.
	As shown in figure \ref{fig:GTSRB}, some images captured suffer from low resolution, poor contrast, brightness, blur, dark light, and tilt motion.
	
	The Belgium Traffic Sign Classification dataset (BTSC) consists of 4533 training images and 2562
	testing images, which is divided into 62 traffic sign
	types. The BTSC dataset is distorted similarly as the GTSRB. As shown in figure \ref{fig:belgiumIssues}, the BTSC dataset suffers from aging, brightness, and perspective issues mainly in the images.
	
	The Croatian Traffic Sign dataset, known as rMASTIF, has 5828 total images: 4044 training images, and
	1784 testing images. It is divided into 31 classes. The
	rMASTIF dataset mainly suffers from aging and blurring
	effects, as shown in figure \ref{fig:rMIssues}.
	
	The GTSRB dataset includes a larger dataset relative to the BTSC dataset, but with less number of different types of traffic signs. As shown in figure \ref{fig:GTSRBdistribution}, the dataset contains 43 unbalanced different types of traffic signs. Compared to the GTSRB dataset, the BTSC dataset is severely unbalanced, as in figure \ref{fig:BTSCdirstibution}. Thus, it increases the difficulty in training and recognition.

	\subsection{Performance Evaluation}
	
	{Our experiments were conducted via Python and Tensorflow framework, running on a laptop with an Intel Core i7-8750H CPU, with a 16GB CPU RAM and an Nvidia Geforce GTX 1070 discrete GPU with 1920 CUDA cores and 8 GB of RAM with a frequency 1.48 GHZ}. Our model as well as our proposed pre-processing technique are GPU accelerated. We use CUPY library  \cite{cupy_learningsys2017} for Numpy and Scipy acceleration on GPU. 
	
	We performed augmentation on the training dataset. All images are resized to 60x60 during pre-processing. To increase the size and variation of the dataset, we applied Keras augmentation on the training dataset with a width-shift range and a height-shift range of 0.1, a 0.2 zoom range, a sheer range of 0.1, and a rotation range of 10 degrees. We split our training dataset into 90\% training, and 10\% validation. 
	
	In all of our experiments, we applied stochastic gradient descent as an optimizer with mini-batches of size 32. We used Keras API Orthogonal for initializing the weights of our model with a gain of 1.0, and applied kernel L2 regularization. Our initial learning rate is 0.01, and we reduce the learning rate with a factor of 0.02.
	
	Our approach achieved the top-1 optimal accuracy of 99.75\% for the GTSRB, out-performing the previous state-of-the-art techniques, and at the same time with fewer memory requirements. We applied a kernel regularizer of 0.01 in all our proposed architecture's convolutional layers for the GTSRB. A Patience of 2 was applied to the reduced learning rate mentioned above. {Our approach was trained for 100 epochs for GTSRB, with 3.2 M parameters. The experiment was repeated 10 times to report the standard deviation in accuracy which came up as 0.1\%.}
	
	We experimented our approach using the BTSC benchmark. Our proposed approach achieved the highest accuracy of 99.25\%, compared to the top-performing techniques. We tuned our proposed model's number of filters. For the first convolutional layer (conv1), 16 filters were used, 32 filters for conv2, 64 filters for conv3, 128 filters for conv4, and for the final convolutional layers (conv5, conv6), 256 filters were used. A kernel regularizer of 0.01 was used for the first 3 convolutional layers (conv1, conv2, conv3), and a kernel regularizer of 0.1 was used for the rest of the convolutional layers(conv4, conv5, conv6). We applied the same reduced learning rate mentioned above but with a patience of 3. Our approach was trained for 125 epochs to reach the optimal accuracy of 99.25\%, with 3.1M parameters.

	\begin{table}[h]
		\caption{Performance comparison for the different architectures of the GTSRB }
		\centering
		\begin{tabular}{|p{2.5cm} |p{1.25cm}| p{1.0cm}| p{1.8cm}|} 
			\hline
			Model Name & Top-1 accuracy & Number of Params & Inference + Pre-processing time\\ 
			\hline
			\ourb & \textbf{99.75}\%& 3.2 M& 0.8-1.3 ms \\

			STDNN \cite{STDNN}  & 99.71\%& 14 M & 4.28 s\\
			LWDNN \cite{Zhang2019LightweightDN} & 99.61\%& 0.8 M & N/A\\
			MCDNN \cite{MCDNN} & 99.46\%& 90 M & N/A  \\
			DHCNN \cite{DHCNN} & 99.33\% & 6.26 M & N/A \\
			COSFIRE \cite{COSFIRE} & 99.25\% & N/A & 17.7 s prepr.\\ 
			OneCNN \cite{OneCNN} & 99.11\% & N/A & 2 ms\\
			HUMAN \cite{HUMAN}& 98.84\% & N/A & N/A \\
			MSCNN \cite{MSCNN}  & 98.31\% & 1.43 M& N/A \\
			Random Forest \cite{MSCNN}  & 96.41\% &N/A& N/A \\
			\hline
		\end{tabular}
		\label{PerformanceComparisonGTSRB}
	\end{table}
	
	\begin{table}[h]
		\caption{Performance comparison for the different architectures of the BTSC }
		\centering
		\begin{tabular}{|c |c |} 
			\hline
			Model Name & Top-1 accuracy \\ 
			\hline
			\ourb & \textbf{99.25}\% \\
			DHCNN \cite{DHCNN} & 99.17\% \\
			LWDNN \cite{Zhang2019LightweightDN} & 99.13\% \\
			STDNN \cite{STDNN}  & 98.86\% \\
			OneCNN \cite{OneCNN} & 98.17\% \\
			\hline
		\end{tabular}
		\label{PerformanceComparisonBTSC}
	\end{table}
	
	\begin{table}[h]
		\caption{Performance comparison for the different architectures of the rMASTIF }
		\centering
		\begin{tabular}{|c |c |} 
			\hline
			Model Name & Top-1 accuracy \\ 
			\hline
			\ourb & \textbf{99.55}\% \\
			OneCNN \cite{OneCNN} & 99.53\% \\
			\hline
		\end{tabular}
		\label{PerformanceComparisonrMASTIF}
	\end{table}
	
	For the rMASTIF benchmark, our approach achieved a 99.55\% recognition accuracy. We used the same proposed architecture as mentioned in section 3.2, but we tuned the model's number of filters for the third convolutional layer (conv3) and the fourth convolutional layer (conv4) to be 64 filters instead of 128. We adjusted the kernel size for the convolutional layers. For conv1, conv3, and conv5, a kernel size of (5x5) is applied; meanwhile, for the rest of the layers, a kernel size (3x3) is applied. A kernel regularizer of 0.1 was applied for all convolutional layers. There were no dropouts used in the model, and patience of 2 was applied to the reduced learning rate mentioned above. Our approach was trained for 50 epochs to reach an optimal accuracy of 99.55\%.
	
	\begin{figure}[h]
		\centering
		\includegraphics[width=\linewidth]{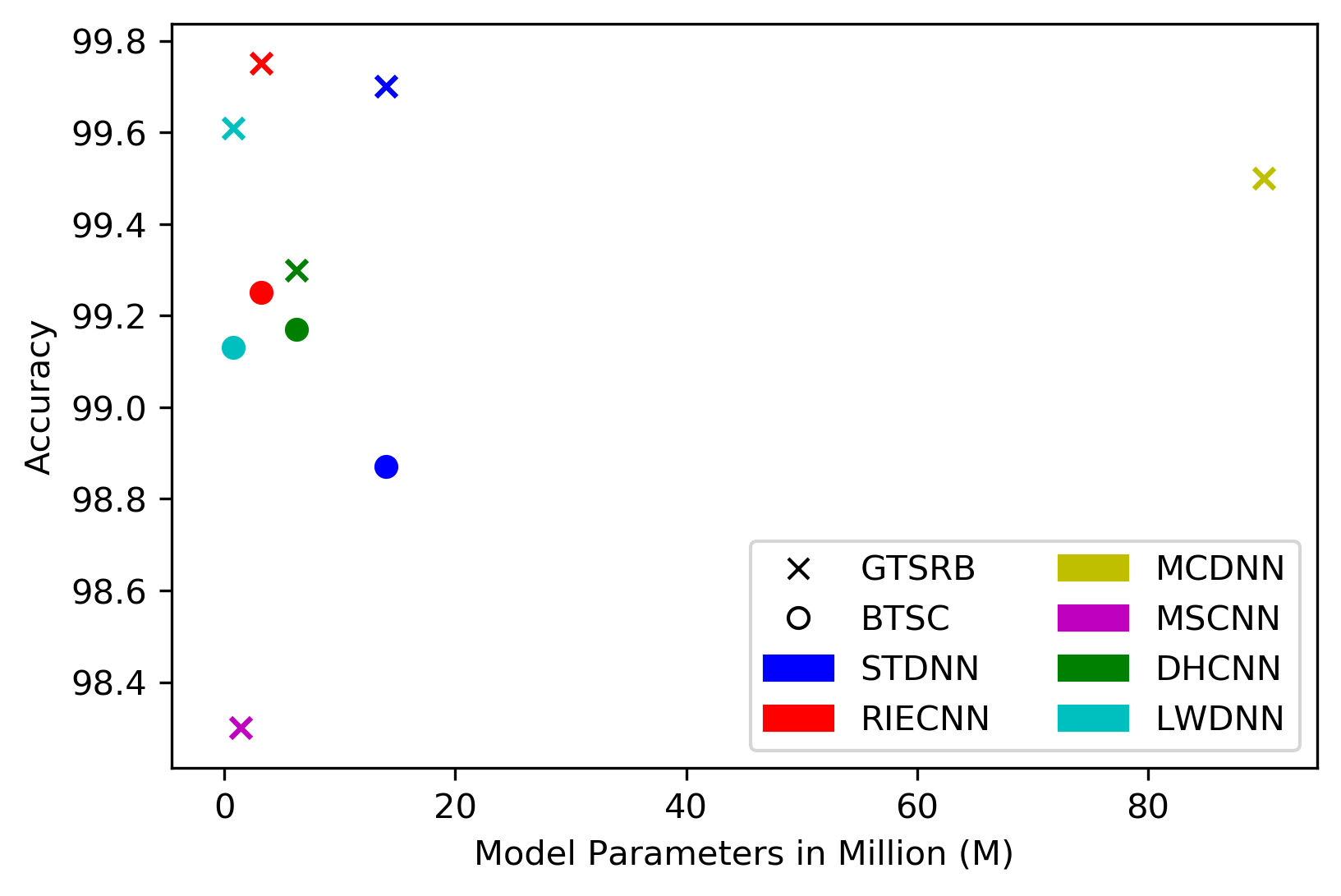}
		\caption{Accuracy vs number of parameters in the GTSRB and the BTSC respectively}
		\label{fig:ParetoDiagramCombined}
	\end{figure}

	Table \ref{PerformanceComparisonGTSRB}
	outlines the top accuracy, number of parameters and inference and pre-processing time of our approach {\our} compared with the state-of-the-art architectures for the GTSRB. We compare our approach's total processing time relative to the rest of techniques' reported total processing time. The machines used to conduct the experiments in each technique including ours are nearly similar specifications. {Our approach takes on average 0.8-1.3 ms per image. Cameras usually record scenes, especially busy scenes with high activity, at 30 fps or 60 fps. Assuming the worst case of 60 fps, there is a 16.67 ms gap between each two consecutive frames. Thus, using our approach will yield an approximately 15 ms outstanding gap that may be used for any further applications.}
	
	Table \ref{PerformanceComparisonGTSRB}, Table \ref{PerformanceComparisonBTSC}, and Table \ref{PerformanceComparisonrMASTIF} show the performance superiority of our approach {\our} compared with the state-of-the-art architectures for the GTSRB, the BTSC, and the rMASTIF respectively. Our approach has achieved an accuracy of 99.75\% on the GTSRB, 99.25\% on the BTSC, and 99.55\% on the rMASTIF.

	\begin{figure}[h]
		\centering
		\includegraphics[width=\linewidth]{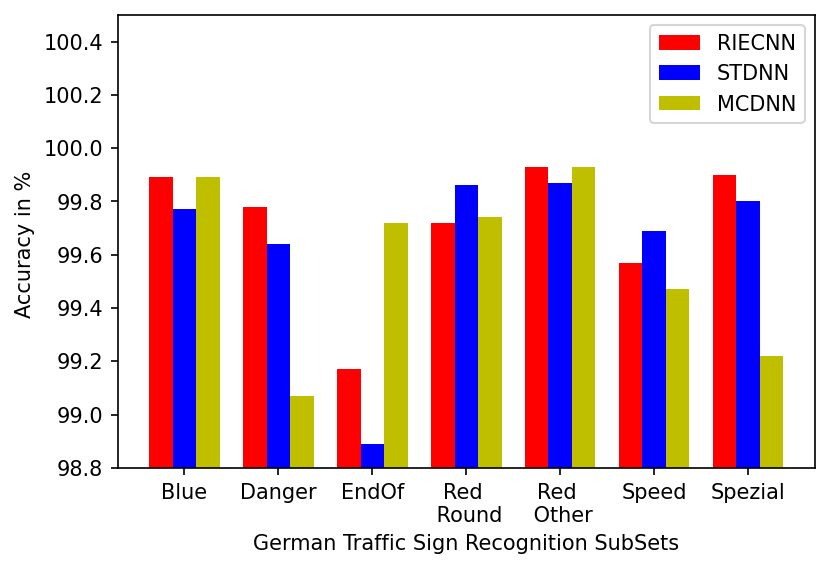}
		\caption{GTSRB 7 signs subsets accuracy comparison}
		\label{fig:SubsetHisto}
	\end{figure} 
	
	\begin{figure}[h]
		\centering
		\includegraphics[width=0.85\linewidth]{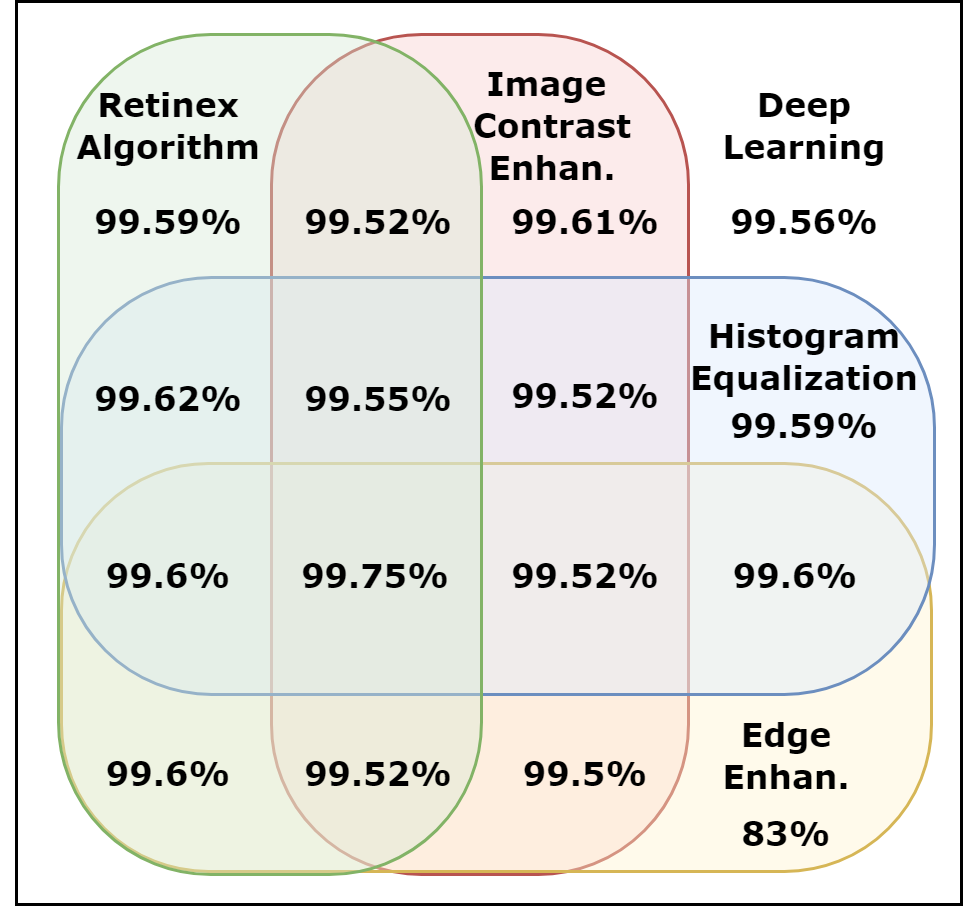}
		\caption{Contribution of each pre-processing method to the performance of the GTSRB}
		\label{fig:Venn}
	\end{figure} 	

	\begin{figure}[h]
		\centering
		\begin{subfigure}[h]{0.2\textwidth}
			\centering
			\includegraphics[height=1.2in]{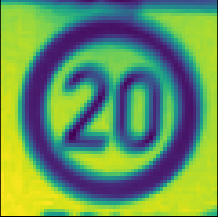}
			\caption{Resulted input image after pre-processing using image contrast enhancement method only}
			\label{YINGImage}
		\end{subfigure}
		~
		\begin{subfigure}[h]{0.2\textwidth}
			\centering
			\includegraphics[height=1.2in]{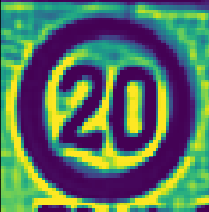}
			\caption{Resulted input image after pre-processing using our proposed 4-stages pre-processing technique}
			\label{AllImage}
		\end{subfigure}%
		\caption{Pre-processed sample image after applying 1-stage versus 4-stages of pre-processing techniques}
		\label{imagetrial}
	\end{figure}

	\begin{figure}[h]
		\centering
		\begin{subfigure}[h]{0.3\textwidth}
			\centering
			\includegraphics[width=\textwidth]{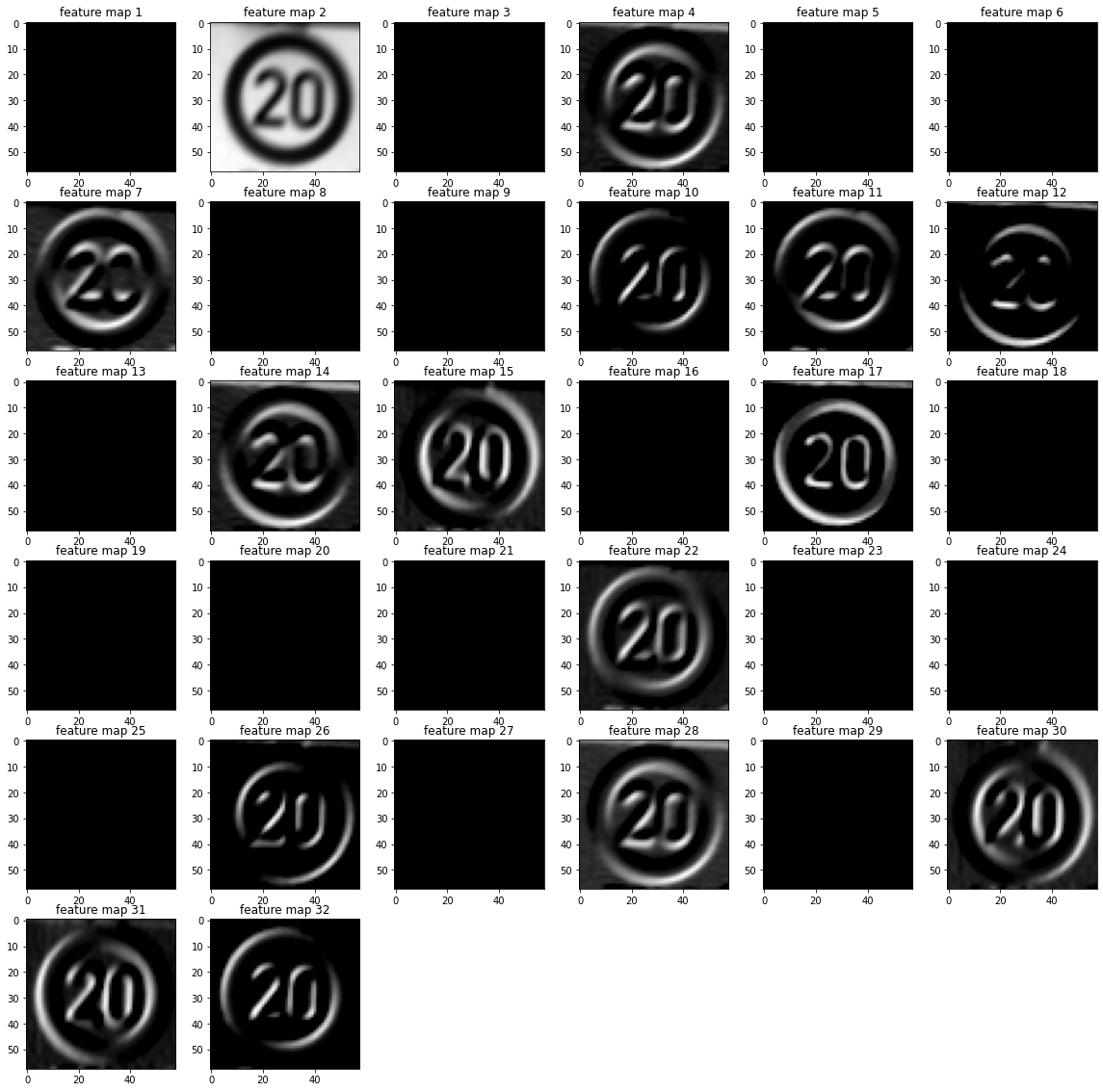}
			\caption{Feature maps for conv1 layer}
		\end{subfigure}%
		
		\begin{subfigure}[h]{0.3\textwidth}
			\centering
			\includegraphics[width=\textwidth]{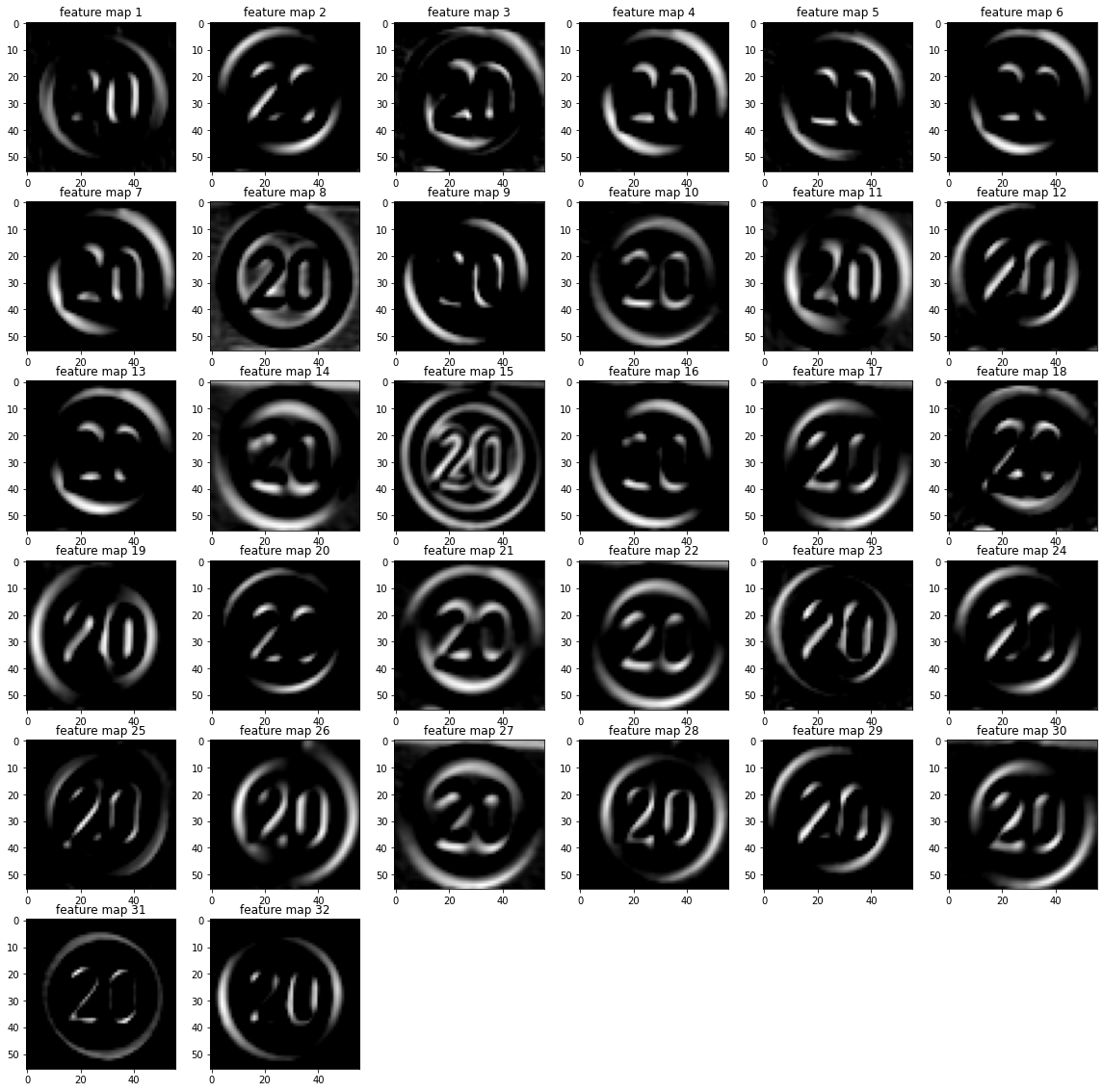}
			\caption{Feature maps for conv2 layer}
		\end{subfigure}
		\caption{Feature maps for figure \ref{YINGImage}}
		\label{featuremapYING}
	\end{figure}

	\begin{figure}[h]
		\centering
		\begin{subfigure}[h]{0.3\textwidth}
			\centering
			\includegraphics[width=\textwidth]{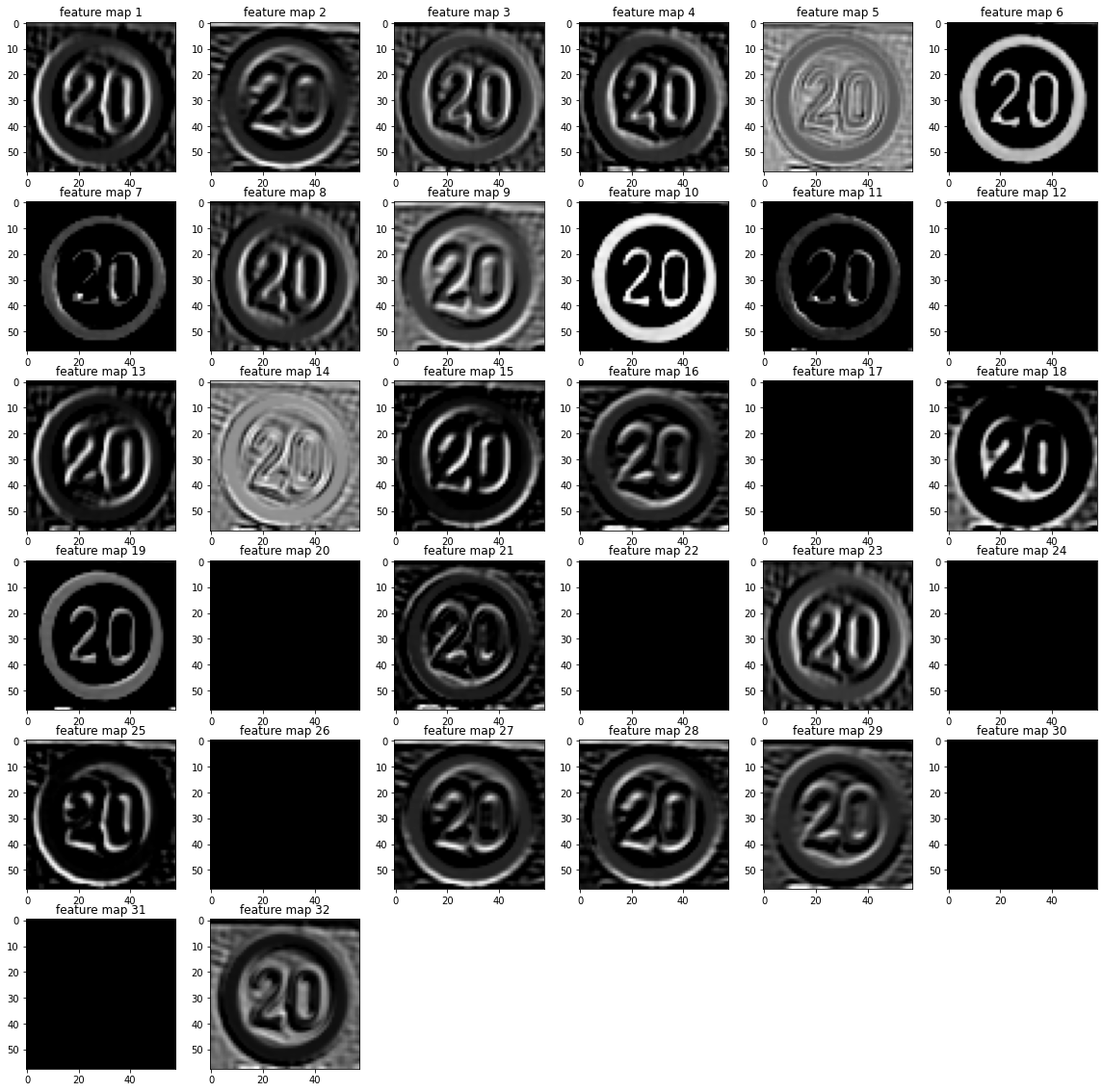}
			\caption{Feature maps for conv1 layer}
		\end{subfigure}%
		
		\begin{subfigure}[h]{0.3\textwidth}
			\centering
			\includegraphics[width=\textwidth]{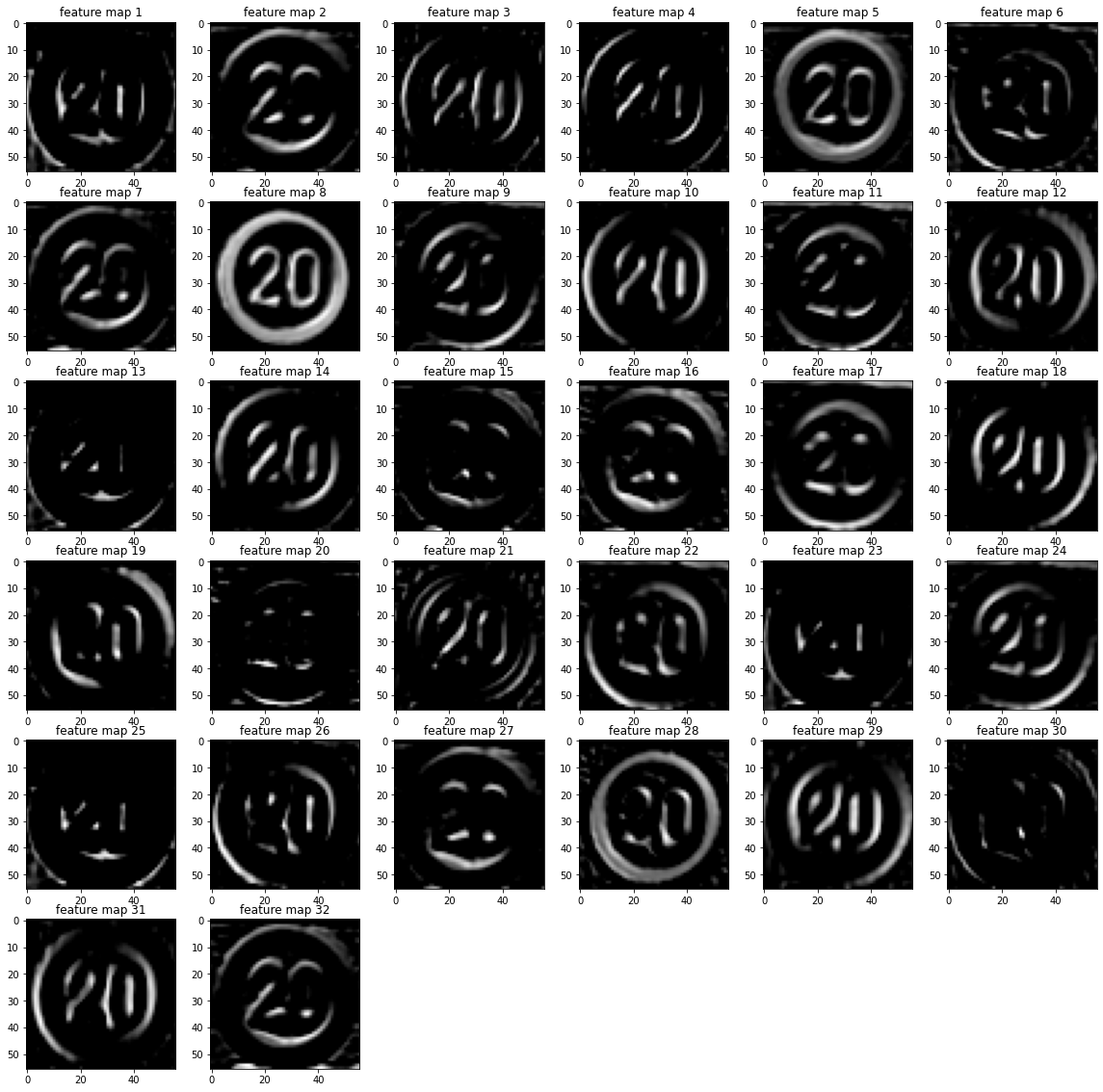}
			\caption{Feature maps for conv2 layer}
		\end{subfigure}
		\caption{Feature maps for figure \ref{AllImage}}
		\label{featuremap}
	\end{figure}
	
	\begin{figure*}[h]
		\centering
		\includegraphics[width=0.85\linewidth]{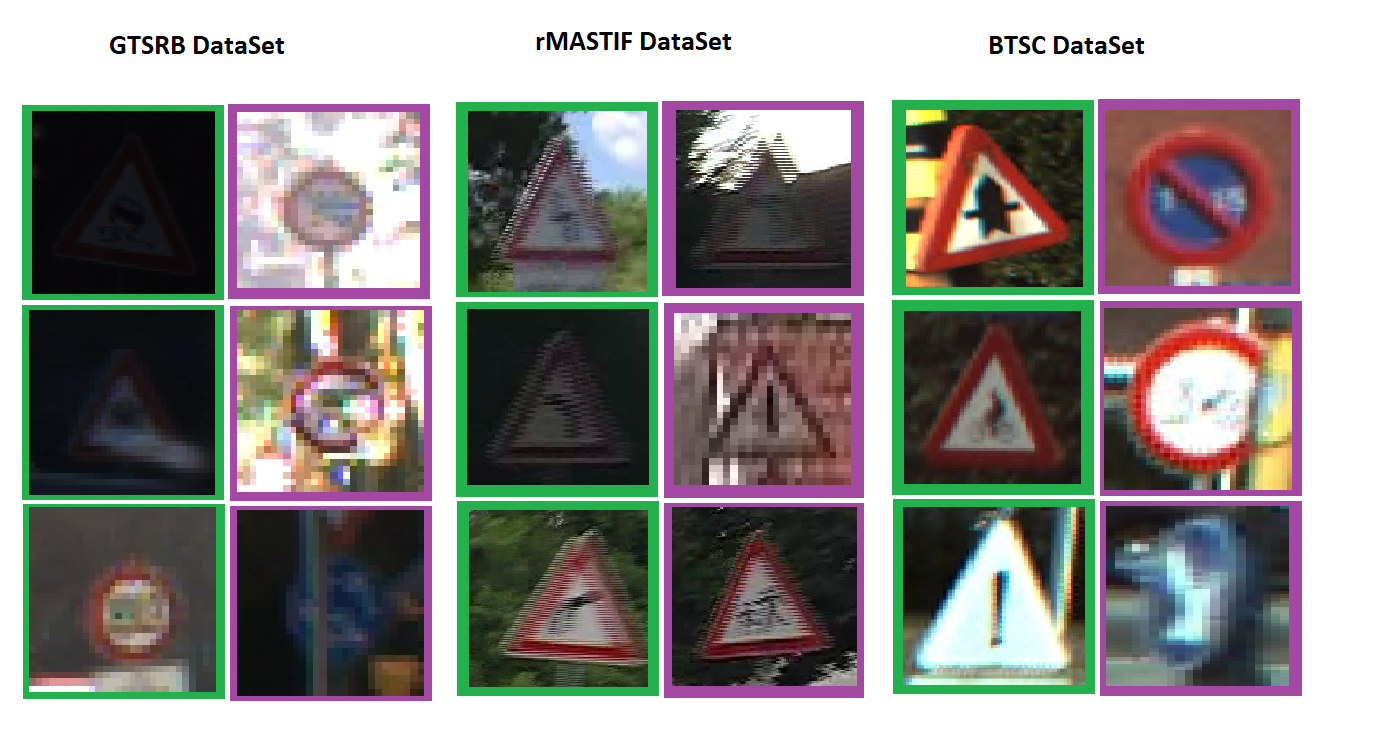}
		\caption{Prediction of RIECNN model on the three datasets, focusing on challenging test set images. Green background indicates correct prediction, purple background indicates miss-classification  }
		\label{fig:pred_inf}
	\end{figure*} 

	Figure \ref{fig:ParetoDiagramCombined} demonstrates a comparison of accuracy versus the number of parameters in the GTSRB and the BTSC respectively. It concludes that our approach achieves the highest-ranked accuracy with 4 times less number of model parameters compared to the STDNN \cite{STDNN} in the GTSRB. For the BTSC, our approach achieves the top-ranked accuracy with 2 times less number of model parameters compared to the DHCNN \cite{DHCNN} in the BTSC.

	Figure \ref{fig:SubsetHisto} shows a comparison of the accuracy of the 7 sign subsets concerning the GTSRB Benchmark for our proposed architecture against the two best-performing architectures. The GTSRB dataset is partitioned into 7 subsets: Blue, Danger, EndOf, RedRound, RedOther, Speed, and Spezial. It highlights that {\our} has demonstrated competitive accuracy in the majority of subsets, and excelling in the Spezial subset. It can be inferred that EndOf and Speed subsets were the subsets with the largest misclassification compared to the other performing architectures.

	\subsection{Observation And Analysis}
	
	Figure \ref{fig:Venn} shows the accuracy of the different combinations of the four stages of our pre-processing method. When combining all pre-processing stages, it yields the best performance with a recognition accuracy of 99.75\%. It can be inferred that using the image contrast enhancement algorithm alone yields 99.61\% which is a competitive accuracy. This concludes that the GTSRB dataset images suffer from contrast and brightness problems mainly.  Moreover, combining the retinex algorithm with histogram equalization yields 99.62\%; meanwhile, applying the retinex algorithm with edge enhancement yields 99.6\% in recognition accuracy. We can conclude that improving the color consistency and enhancing edges in the images helps to improve the resolution for poor-quality and blurry images.

	We investigated the impact of our pre-processing technique on our proposed CNN architecture's first convolutional (conv1) and second convolutional (conv2) layer feature maps. We conducted two experiments on a sample 20-speed limit image. In the first experiment, we applied only the image contrast enhancement stage on that sample image. The resulting image is shown in figure \ref{YINGImage}. The second experiment is using our proposed 4-stages pre-processing technique. The resulting image is shown in \ref{AllImage}. Figure \ref{featuremapYING} shows the conv1 and conv2 layer output feature maps using only the image contrast enhancement stage; meanwhile, figure \ref{featuremap} shows the conv1 and conv2 output feature maps using our proposed 4-stages pre-processing technique. It can be inferred from both figures that the conv1 layer focuses on shape, content; meanwhile, the conv2 layer focuses more on edges, invariant features. Compared to figure \ref{featuremapYING}, it can be concluded from figure \ref{featuremap} that the feature maps highlight finer distinctive features with more detail. The outputs of the feature maps show different variations in the features. We believe that it helps the model learn to focus more on the distinctive features invariant to transformations which results in better classification.

	Figure \ref{fig:pred_inf} shows the prediction of RIECNN model on challenging test images in GTSRB, rMASTIF, BTSC benchmarks. The green background indicates correct classification, while purple background indicates miss-classification. RIECNN can predict challenging images correctly under poor and diverse environmental conditions, and it can be observed that the misclassified images can be very difficult to classify by the human eye due to poor resolution. As shown, RIECNN model handles most of challenges opposed by GTSRB, rMASTIF, and BTSC datasets. 
	
	\section{Conclusion}
	In this study, we present our novel approach Real-Time Image Enhanced CNN (RIECNN) for traffic sign recognition. Our methodology is divided in two stages. First, captured images are pre-processed to enhance images' quality, and then the images are fed to our deep learning CNN architecture. It was shown that the Retinex Algorithm combined with image enhancement algorithms in the pre-processing stage contribute together to yield competitive recognition accuracy. We experimented our approach RIECNN with multiple datasets: the GTSRB, the BTSC, and the rMASTIF. RIECNN demonstrates a strong generalization ability with the highest recognition accuracy and much fewer parameters than previous techniques. Our approach achieves the highest rank in the GTSRB, the BTSC, and rMASTIF out-performing all previous state-of-the-art techniques with a 99.75\% recognition accuracy in the GTSRB, 99.25\% in the BTSC and 99.55\% accuracy in the rMASTIF. 
	
	The limitation in our approach mainly resides that the recognition accuracy of the EndOf and Speed subsets in the GTSRB benchmark is relatively lower, compared to other state-of-the-art architectures. Meanwhile, the misclassified images by our approach in these subsets can be yet still pretty difficult for humans to classify. Future work will focus to experiment our approach on other publicly available traffic sign datasets and investigate our approach's robustness. Also, further enhancements could be applied to the model or pre-processing stage to boost our approach's accuracy and reduce misclassification rates of specific subsets in the GTSRB.

%Bibliography
\bibliographystyle{unsrt}  
\bibliography{references}

\end{document}